# Influence of image noise on crack detection performance of deep convolutional neural networks

R. Chianese[1], A. Nguyen[1], V.R Gharehbaghi[2], T. Aravinthan[1], M. Noori[3]
[1]Department of Civil Eng., Faculty of Health, Eng. & Sciences, University of Southern Queensland, Qld. Australia
[2]Department of Civil Eng., Faculty of Engineering, Kharazmi University (Tehran) Iran
[3]Department of Mechanical Eng., Faculty of Engineering, California Polytechnic State University, San Luis Obispo California, United States
email: u1083577@umail.usq.edu.au, andy.nguyen@usq.edu.au (corresponding author), vahidrqa@gmail.com, thiru.aravinthan@usq.edu.au, mnoori@calpoly.edu

ABSTRACT: Development of deep learning techniques to analyse image data is an expansive and emerging field. The benefits of tracking, identifying, measuring, and sorting features of interest from image data has endless applications for saving cost, time, and improving safety. Much research has been conducted on classifying cracks from image data using deep convolutional neural networks; however, minimal research has been conducted to study the efficacy of network performance when noisy images are used. This paper will address the problem and is dedicated to investigating the influence of image noise on network accuracy. The methods used incorporate a benchmark image data set, which is purposely deteriorated with two types of noise, followed by treatment with image enhancement pre-processing techniques. These images, including their native counterparts, are then used to train and validate two different networks to study the differences in accuracy and performance. Results from this research reveal that noisy images have a moderate to high impact on the network's capability to accurately classify images despite the application of image pre-processing. A new index has been developed for finding the most efficient method for classification in terms of computation timing and accuracy. Consequently, AlexNet was selected as the most efficient model based on the proposed index.

## 1 INTRODUCTION

Cracks develop in metal and concrete structures due to many reasons, for example, cyclic fatigue loading, stress corrosion, seasonal temperature fluctuations, mechanical damage, material aging, Alkali-Silica reaction, welding operations, and during manufacturing. Cracks that are not detected and repaired are likely to lengthen and deepen resulting in eventual structural catastrophic failure in some cases.

Tunnels and bridges are examples of civil structures that can cause loss of life or serious asset damage should they fail. The size, height, location and access constraints with these structures can make identification of cracks dangerous and time-consuming [1, 2]. Islam and Jong-Myon [3] acknowledge that visual inspections of large civil infrastructure is a common trend for maintaining their reliability and structural health. Other than safety and time inefficiencies associated with manual crack detection methods, Kim, Ahn [4] advises that manual visual inspections are often considered to be ineffective with regards to accuracy, reliability and cost.

It is well understood that identification of cracks in common civil materials such as concrete and metals is vital; however, the trend to conduct manual visual inspections remains a commonly adopted practice that has limitations [5]. There are safety risks associated with conducting manual inspections and there are even more severe safety risks with not embarking on any structural health monitoring (SHM) regime [6]. The more recent Morandi bridge failure in Genoa Italy, is an example of structural failure due to questionable maintenance practice and high-cost constraints [7].

Detection of cracks in concrete materials has largely been conducted by visual inspections which is known to be ineffective but time-consuming. The recent application of computer vision and deep convolution neural networks (DCNNs) has become an emerging subfield for crack detection, which has made a contribution to improved safety, reduced cost and time [8, 9]. Computer vision is a type of artificial intelligence that trains computers to identify and classify objects from images [10].

The input images used in computer vision method are captured by various means. Unmanned aerial vehicles (UAVs) are becoming a popular choice for high-rise building façade visual crack detection, according to Liu et al. [11]. Images can also be taken by manual means by handheld cameras or cameras can be transported by vehicles and robotic methods such as the methods utilized by Chen et al. [12]. Image capture is subject to the limitations of noise, which can originate from the image capture equipment or relative translations between image capture equipment and the image subject [13]. Such motion noise is a common error when image capture equipment is mounted to UAV equipment [14].

Despite the growing use of DCNN for crack classification applications, there are some remaining problems to be addressed. Little research has been conducted on the impact of using lower quality images to detect cracks or how to overcome those effects through the use of image pre-processing tools. The lack of research investment in this area has created uncertainty when noisy images are presented for SHM applications. The significance of this research is the bridging of this uncertainty when comparing the performance of different techniques.

Referring to the above preface, the outline of this article is as follows: Firstly, some of the related papers in the realm of the impact of noise and computer vision are reviewed. Following this, different types of noises are introduced and the image data set is presented, including the introduction of the proposed methodology for comparing image enhancement techniques. Finally, the results for each technique are depicted and discussed in detail.

## 2 RELATED WORK

The use of image data in computer vision-based crack detection has recently become a hot topic in the non-contact monitoring approach as opposed to the contact type health monitoring such as the successful methods deployed by researchers Nguyen et al., whereby FE modeling and model updating exercises are performed based on inputs obtained from single and triaxial sensors affixed to structures [15, 16]. Similarly, structural deterioration and damage can also be classified from contact sensor data using signal processing techniques and supervised machine learning approaches as recently introduced by Gharehbaghi et al. [17, 18].

In the area of computer vision-based crack detection, deep learning algorithms are used to classify cracks from non-contact means such as image data. The two main approaches in this direction are (i) deep learning by pre-trained models (aka transfer learning) and (ii) deep learning based on new model developments. The advantage of pre-trained models is the reduction in time required to gain skills and to develop alternative model network architectures. Depending on the application, pre-trained models can also speed up the training process [19]. Pre-trained models require images as input data that need to be resized, which results in interpolation between image features in the time-frequency space, which has a consequence of reduced network accuracy.

Notable research utilizing pre-trained models include that conducted by Bang et al. [20], who compared VGG-16 and ResNet-101 against their own designed network tasked at detecting pavement cracks. Results reveal that pre-trained networks had lesser accuracy compared to the researcher's approach with their own black-box asphalt image set. Similarly, Cho and Kim[21] used AlexNet to train images scraped from the internet and their own UAV images to classify cracks, joints, edges and other surface irregularities in concrete materials. A probability map was generated; however, many images contained very well-defined surface irregularities with little noise impact. Other pre-trained networks such as VGG-16, InceptionV3 and Resnet have been tasked with classifying concrete crack images by automated means with well-defined concrete cracks [22]. The comparison of networks was, however, not completed on noisy images. AlexNet, InceptionV3, and Resnet pre-trained models have also been compared, utilizing well-defined crack images with no discernible image noise [3].

Currently, there is limited research investigating the impact of image blur and Salt & Pepper noise incorporating automated vision-based crack detection utilizing pre-trained methods. This study will bridge this gap and furthermore, install confidence with the use of poor image quality for classification tasks.

## 3 NOISE ON IMAGES AND DATASETS

Image noise occurs due to image acquisition and image transmission. The main types of noise include impulse, Gaussian, Rayleigh, Erlang, Exponential, Uniform and Impulse (salt & pepper) [23]. Impulse noise can be caused by incorrect ISO digital settings [24]. Gaussian noise can occur due to elevated image sensor temperatures and due to voltage and illumination variables [25]. Blur noise is also a type of image acquisition noise caused by the relative movement between the image subject and acquisition equipment. For computer vision and AI image processing with CNNs, image noise poses as a threat to classification accuracy performance.

Popular public image datasets for the development of SHM crack detection techniques include SDNET-2018 [26], which contains over 56 000 images, Mendeley data with 40 000 images [27] and GoogleNet [28] with a growing image database. This research will use SDNET-2018 due to its challenging images, which includes a large variety of crack image width and sizes, including concrete surface irregularities and minor cracks. Samples of the less challenging Mendeley images can be viewed in Figure 1 below, with images A-D showing cracks and the bottom row E-H without cracks.

Similarly, sample challenging SDNET-2018 images used in this research can be viewed in figure 2 below, with the top row (A-D) showing cracks. Due to the thin and lesser defined crack images, the data is more challenging to classify.

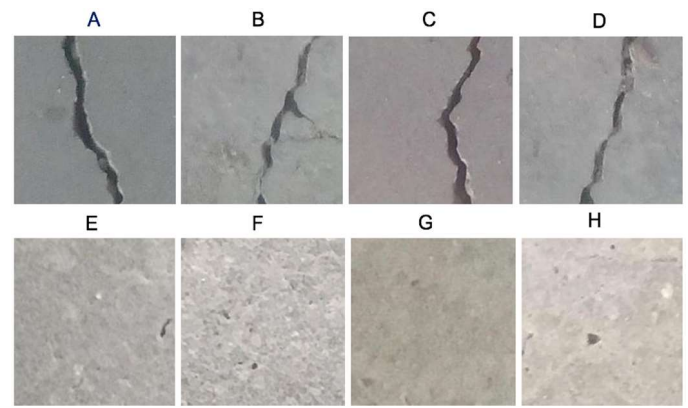

Figure 1 Mendeley data

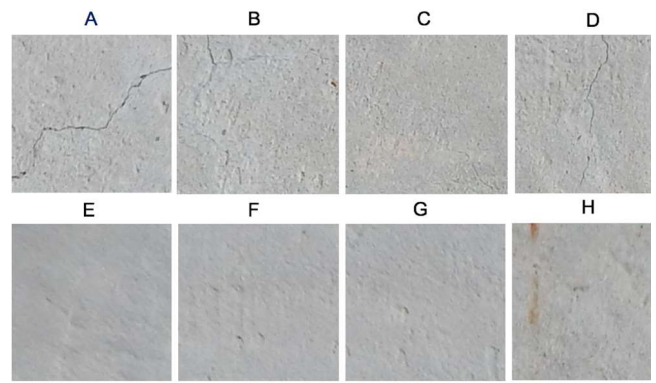

Figure 2 SDNET-2018

Figure 3 shows samples of SDNET-2018 dataset in the existence of motion blur and salt & pepper noise. As can be seen, identifying minor cracks in a noisy environment is a changeling issue. Therefore this image data is selected for comparing the performance of different methods in classifying the crack and non-crack image. In the following, the proposed methodology for this target is described.

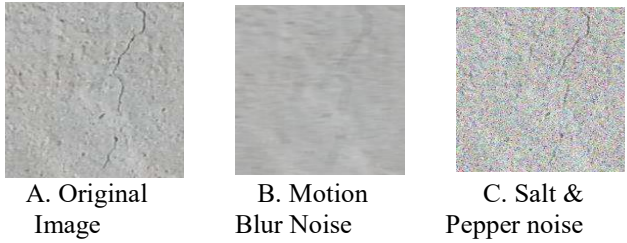

A. Original Image   B. Motion Blur Noise   C. Salt & Pepper noise

Figure 3 Noise impact on SDNET-2018

## 4 METHODOLOGY

In this study, images sourced from SDNET-2018 are strategically processed based on the workflow depicted in figure 4 below. Firstly two types of noise, including blur and salt and pepper, are applied to the image separately. Afterward, three methods include M1 (pure transfer learning), M2 (transfer learning coupled with 2d-adaptive noise filtering), and M3 (transfer learning coupled with image sharpening), are deployed for classifying crack and non-crack images. Finally, some indices are practices for assessing image classification performance.

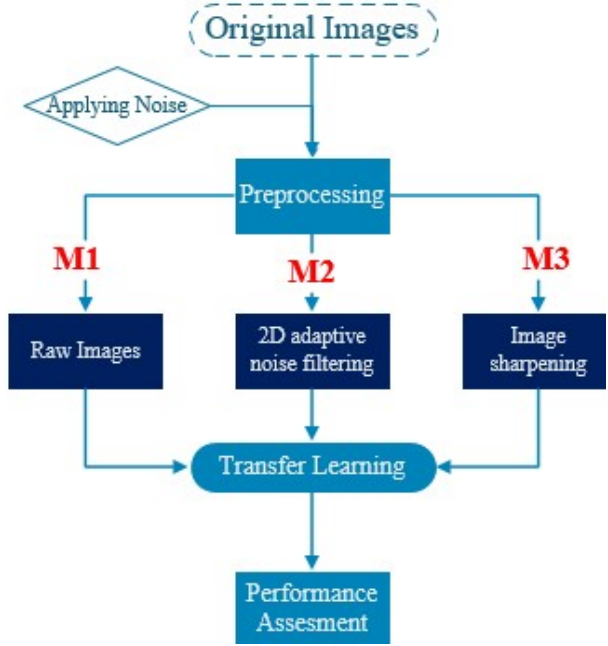

Figure 4 Research workflow

For the purpose of using a pre-trained model, the input layer should be modified according to the new image data. Therefore, in the first step, the original size of images is changed to the size of the model. Image pre-processing includes the resizing of SDNET-2018 RGB images from 256×256×3 to 224×224×3 for Resnet-101, AlexNet requires image input size of 227×227 and Inception-V3 requires 229×229×3. All images are normalized after resizing and applying filtering techniques to ensure the image space is centralized about the image centroid.

Regarding method M1, no further processing is applied to raw images, which are then used to refine the transfer learning CNN hyperparameters for enhanced performance.

### 4.1 Image filtering methods (M2 and M3):

In the following approaches, two filtering techniques are applied to the noisy images to test network efficacy in the pre-trained networks: 2D adaptive noise Wiener filtering and image sharpening. 2D adaptive noise filter calculates the local mean and variance around each pixel as follows [29, 30]:

$$\delta = \frac{1}{PQ} \sum_{p_1,p_2 \in \delta} s(p_1, p_2) \quad (1)$$

$$\sigma^2 = \frac{1}{PQ} \sum_{p_1,p_2 \in \delta} s^2(p_1, p_2) - \delta^2 \quad (2)$$

Where $\delta$ denotes a $P \times Q$ local neighbourhood of each pixel in a sample image, s being the power spectrum with $p_1$ and $p_2$ being the upper and lower frequency bands.

Next, a filter is created by using the average of the local estimated variances:

$$r(p_1, p_2) = \delta + \frac{\sigma^2 - v^2}{\sigma^2}(s(p_1, p_2) - \delta) \quad (3)$$

Where v2 is the average of the noise variance, and when it is not available, the adaptive filter deploys the mean of all local variances instead.

In method M3, the image sharpening technique used is called unsharp masking (USM) sharpening for enhancing edge contrast. The USM sharpening approach contains the following steps:

Firstly, Gaussian high-pass filtering is applied as follows[31]:

$$H(x,y) = I_i(x,y) - I_i(x,y) \otimes G_\sigma \quad (4)$$

where $H$ presents the high pass filter for the original image $I_i(x,y)$ stands for coordinates, and $G_\sigma$ is Gaussian high-pass filter. Herein, the standard deviation $\sigma$ controls the sharping range.

Secondly, an unsharpened mask is added to the original image as:

$$O(x,y) = I_i(x,y) + \lambda H(x,y) \quad (5)$$

where $O(x,y)$ presents the sharpened image and $\lambda$ the sharpening.

### 4.2 Transfer learning

Most popular transfer learning methods are based on convolutional neural networks (CNNs). CNNs are one of the main subsets of deep neural networks with artificial neurons deployed for image classification and recognition. A typical CNN includes a set of convolution and pooling layers as a feature extraction part and fully connected layers and activation function as classification part for classification objects. A typical CNN architecture is shown in Figure 5 [32]. In transfer learning, the CNN, which is previously pre-trained with a significant number of images, is deployed for the classification of a new dataset with a modified classification part. Subsequently, the pre-trained models are able to transfer the previous knowledge such as weights and features gained from one task (source task) is reused to a new, however similar task (target task) with less computational resources and training time [33].

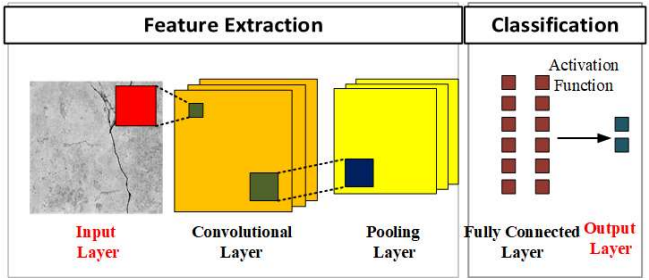

Figure 5 Typical CNN architecture

There are several pre-trained deep learning models, including Xception, VGG16, Vgg19, ResNet50, and MobileNet have been used for image classification. However, in this study, three pre-trained models, including AlexNet, Inception-V3 and ResNet101, have been repurposed through a process of transfer learning. These models are selected so as to conduct a comparison between models with various model complexity as well as computational complexity. For illustration purposes, a ball chart presentation of different models taken from Top-1 accuracy on the ImageNet-1k is reported in Figure 6. In this figure, each model's computational cost is considered through the floating-point operations (FLOPs), and the area of each ball presents the model complexity (total amount of learnable parameters) [34].

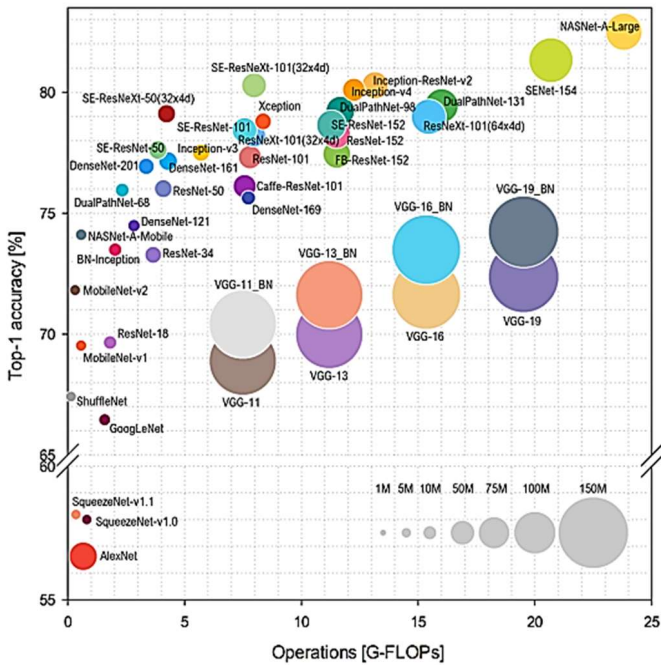

Figure 6 Ball chart reporting the Top-1 accuracy versus computational complexity [34]

Referring to the figure, AlexNet is a candidate for rapid models with fewer operations required but not very accurate; and Inception-V3 and ResNet101 are amongst the top models for accuracy with higher computational complexity (higher amount of parameters). As an example case of network repurposing, Figure 7 shows the modified AlexNet layers. These networks were previously trained using for than 1 000 000 images from the ImageNet database to classify more than 1000 image classes [35]. CNN model repurposing includes modification of the network input and final layers to classify cracked and non-cracked concrete images from SDNET-2018 data. Similarly, Inception-V3 and ResNet-101 network models have their input and last fully connected layers modified to suit the repurposed application to classify cracks from concrete images. Each model is trained with 1000 SDNET-2018 images containing cracks and non-cracked base image data.

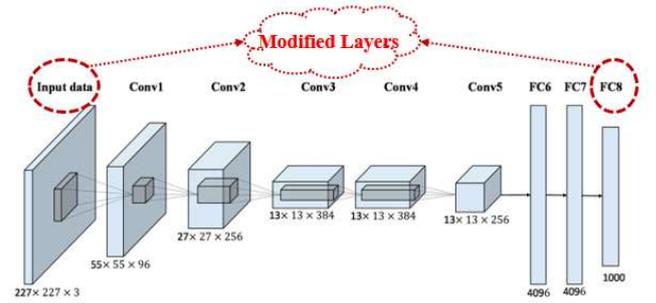

Figure 7. AlexNet modified layers (adapted from [36])

### 4.3 Performance indices

Network performance is measured by means of various indicators such as precision, recall, and F1 Score. Precision is the percentage of results that are relevant [37]. Recall is the percentage of total relevant results that have been correctly classified [37]; recall is also known as Validation accuracy. F1 Score is a combination of precision and recall measures synthesized into a single and convenient measure for DCNN performance comparisons. These indices are formulated as follows[37]:

$$\text{Precision} = \frac{\text{True Positive}}{\text{Actual Results}} \quad (6)$$

$$\text{Recall} = \frac{\text{True Positive}}{\text{Predicted Results}} \quad (7)$$

$$\text{F1Score} = \frac{2 \times \text{Precision}}{\text{Precision} + \text{Recall}} \quad (8)$$

In this paper, for the purpose of having a union comparison, a new comparative performance index (CI) is established as the average of the three above indices.

$$\text{CI} = \frac{\text{Precision} + \text{Recall} + \text{F1Score}}{3} \quad (9)$$

## 5 RESULTS AND DISCUSSION

In this section, the results are demonstrated for three methods (M1, M2 and M3) separately. To show the efficiency of each technique, the percentage of improvement is reported versus the reference benchmarks mentioned in Table 1. This table consists of the performance results of classification using transfer learning applied to original images (noise-free). These attribute results are utilized as reference benchmarks for each model, i.e. Ref.1 for AlexNet, Ref.2 for Inception-V3, and Ref.3 for ResNet101.

Table 1. Baseline network results - no noise (References)

| Attribute % | AlexNet (Ref.1) | Inception-V3 (Ref.2) | ResNet-101 (Ref.3) |
|---|---|---|---|
| Validation | 87.30 | 84.67 | 86.00 |
| Precision | 87.34 | 84.72 | 87.30 |
| F1 | 87.34 | 84.69 | 86.64 |

Furthermore, in order to have the same condition for pre-trained models, the hyperparameters for training is set as depicted in Table 2. It noteworthy that the best hyperparameters can be obtained exclusively for each model through an optimization solver like the Bayesian algorithm. The system configuration used for analysis is shown in Table 3.

Table 2. Hyper Parameters

| Training Solver | Stochastic Gradient Descent with Momentum (SGDM) |
|---|---|
| Initial Learn Rate | 0.0001 |
| Minimum Batch Size | 16 |
| Validation Frequency | 5 |
| Maximum Epochs | 10 |

Table 3. System configuration

| GPU | GTX 1050 | 4 GB/GDDR5 |
|---|---|---|
| RAM | DDR4 | 8 GB |
| CPU | Ryzen5-3500 | 4 Cores/8 Threads |

### 5.1 M1 (Transfer Learning Method)

Applying the first method (M1) the results for each type of noise is shown in Figure 8 and 9. It is evident that the performance is reduced in all models due to noise. Referring to Figure 8, the most efficient model for salt and pepper noise is Inception-V3 with about 78.0%, and the worst case is ResNet-101 with approximately 70.0% accuracy. Additionally, the maximum variation observed in the performance was for ResNet-101 with an observed 16.5% change. This result depicts that ResNet-101 is the most sensitive model to the salt and pepper noise.

By observing Figure 9, AlexNet reveals better performance for motion blur noise with about 82%, and ResNet-101 has the worst performance of around 78.0% accuracy. Comparing both figures, it is deducted that salt and pepper noise has more impact on decreasing the classification accuracy using transfer learning.

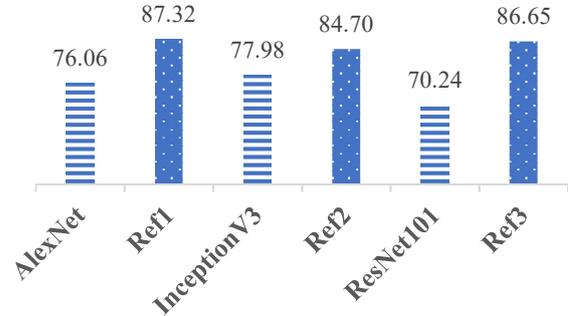

Figure 8 Results for M1 (Salt and Pepper noise)

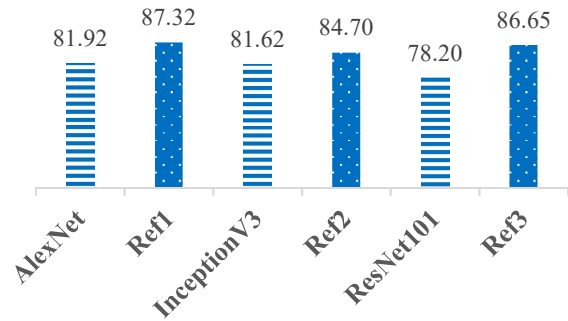

Figure 9 Results for M1 (Motion blur noise)

### 5.2 M2 (2d adaptive noise filtering)

With regards to the second approach and studying Figure 10 and 11, it is evident that both types of noises reveal a negative impact on the performance of classification. In the case of salt and pepper noise, AlexNet and Inception-V3 are suitable candidates for classification with a performance of about 78.5%; however, ResNet-101 is not an appropriate model for comparison purposes. The most significant reduction in performance is observed for ResNet101, which is almost 13%.

Contrarily, AlexNet, and Inception-V3, are not appropriate alternatives for motion blue noise since they experience more than a 10% fall in performance level. Nonetheless, ResNet-101 is the most efficient model, which indicates a nearly 5% drop in classification accuracy.

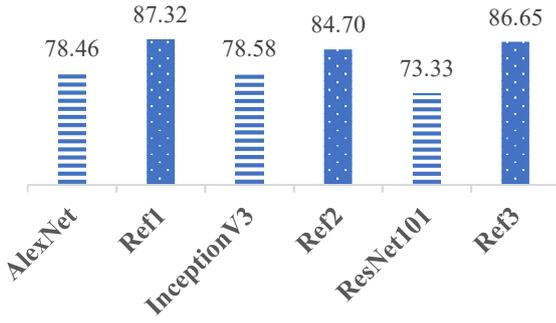

Figure 10 Results for M2 (Salt and Pepper noise)

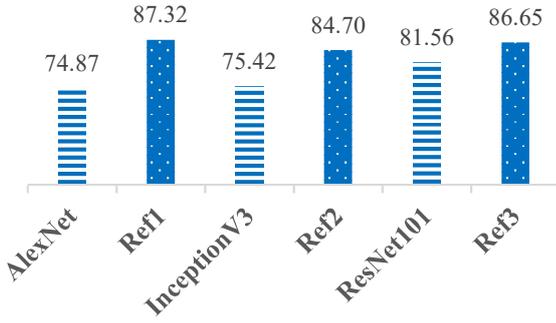

Figure 11 Results for M2 (Motion blur noise)

### 5.3 M3 (Image sharpening)

The classification results for the last approach are depicted in Figure 12 and 13, where the maximum performance of approximately 78.2% is achieved by ResNet-101 and the minimum by Inception-V3 with 69.42% in the existence of salt and pepper noise.

When motion blur noise is applied to images, the top performance is observed for ResNet101 with roughly 86.0%. Inception-V3 reveals relatively lower functionality in this case, with about 82.0% accuracy. Considering both noises, salt and pepper noise discloses considerably higher effects on crack classification performance.

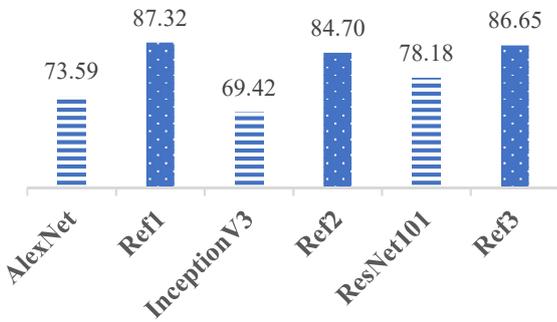

Figure 12 Results for M3 (Salt and Pepper noise)

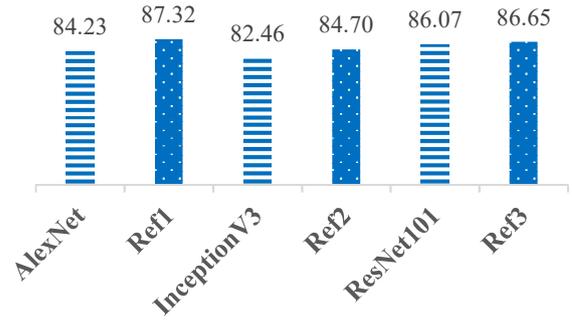

Figure 13 Results for M3 (Motion blur noise)

### 5.4 Efficiency index

For the purpose of comparing the computational time in running each method, the average elapsed time in minutes is calculated for all methods and presented separately for each model (Figure 14). It is clearly evident that AlexNet is more than five times faster than the other two models. Inception-V3 results in a computational time of approximately 27 minutes for processing the test images. The most computationally complex deep learning model is Resnet101, which requires approximately 33.0 minutes for completing a classification process.

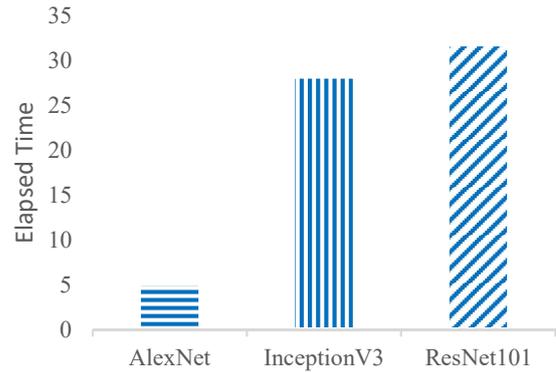

Figure 14 Computational time assessment

Finally, for the sake of determining the most cost-effective solution for the noise environment, computational efficiency criteria (Cc) is established such that it can nominate the most accurate method while it demands less computational time. This evaluation criterion is defined through an index as:

$$C_{c,i} = \frac{1+CI_i}{(1+CI_i) \times C_{c,\max}} \quad \& \quad CT \geq 1.0 \tag{10}$$

Where $CI_i$ stands for comparative performance index for the $i^{th}$ method and $CT_i$ is the computation time in minutes calculated for each method. $C_{c,\max}$ is the maximum value of $CI_i$ which is used for normalization. In this equation, the minimum computation time is supposed to be one minute. Ignoring the normalization parameter, $C_{c,\max}$, the highest limit of $C_c$ would be 100, which is derived from $CI = 1.0$ and $CT = 1.0$. Contrarily, the lowest limit of this index is zero deriving from $CI = 0.0$ and $CT \to \infty$.

Values near one indicate the superior, cost-effective method, while the lower values denote the more costly method. The results for each method and each type of noise is indicated in Figure 15 and 16. As highlighted, the proposed index is not sensitive to the type of image and takes the same values for both types of noise. Therefore, despite the fact that the AlexNet was not an accurate model in all cases, in respect of computational cost criteria, it is the leading model in all methods. Likewise, InceptionV3 and ResNet101 depict nearly the same functionality based on Cc criteria.

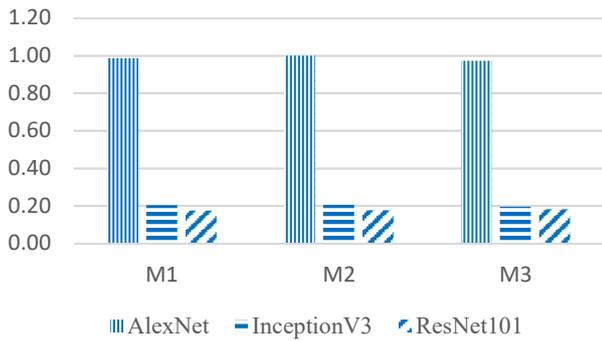

Figure 15 Computational efficiency index Cc (Salt and Pepper noise)

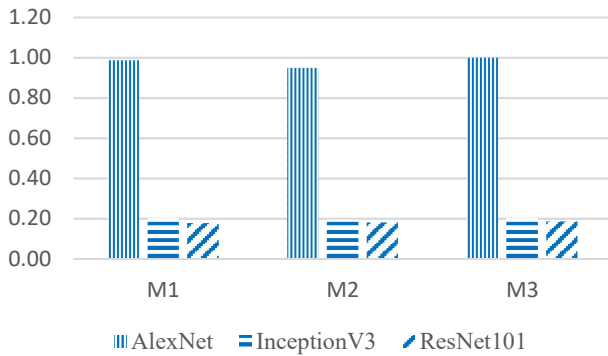

Figure 16 Computational efficiency index Cc (Motion blur noise)

## 6 CONCLUSION

This research successfully evaluates the influence of image blur noise on AlexNet, Inception-V3, and ResNet-101 CNN accuracy. Three different approaches, including transfer learning without image enhancement, transfer learning coupled with 2d-adaptive noise filtering and image sharpening, were utilized for assessing the effect of different noise on the performance of image classification.

As a result, it was shown that 2d-adaptive image filtering applied with transfer learning (M2) is the best solution for addressing salt and paper noises. Whereas, noise mitigation by means of image sharpening (M3) was the most appropriate method for dealing with motion blur image noise. As a summary of results, network performance improvements for these two effective methods and corresponding models (AlexNet, Inception-V3, and Resnet-101) are shown in Table 4 below.

Table 4. Appropriate Noise mitigation techniques

| Method | Model | Performance (%) | Suitable state |
|---|---|---|---|
| M2 (2d adaptive noise filtering) | AlexNet & Inception-V3 | 78.5% | Salt and pepper noise |
| M3 (Image sharpening) | ResNet-101 | 86.0% | Motion Blur noise |

In order to assess the efficiency of each method and model, a criterion called Cc was developed, which considered both accuracy and computational time simultaneously. The proposed index revealed that AlexNet was the best CNN for all considered methods concerning overall performance.

Besides the excellent advantages of transfer learning for image classification applications, such as classifying concrete images, it has been found that there is a moderate to high impact on classification accuracy when blur noise or Salt & Pepper is introduced to the images. As an extension of this research, new approaches utilizing non-transfer learning techniques, including signal transformation and feature conditioning coupled with hyper-parameters optimization techniques, is underway to help with coping with image data uncertainties such as noise. Additionally, testing of these techniques with other types of noise, such as Impulse, Erlang, and Rayleigh noise, presents a recommended future research opportunity. This and the extended research have opened a number of opportunities for future research on deep learning for structural health monitoring of civil infrastructure.

## 7 REFERENCES

1. Gang, L., et al. Automatic Tunnel Crack Detection Based on U-Net and a Convolutional Neural Network with Alternately Updated Clique. Sensors, 2020. 20, 717 DOI: 10.3390/s20030717.
2. Song, Q., et al. Real-Time Tunnel Crack Analysis System via Deep Learning. IEEE Access, 2019. 7, 64186-64197 DOI: 10.1109/ACCESS.2019.2916330.
3. Islam, M.M.M. and K. Jong-Myon Vision-Based Autonomous Crack Detection of Concrete Structures Using a Fully Convolutional Encoder–Decoder Network. Sensors, 2019. 19, 4251 DOI: 10.3390/s19194251.
4. Kim, H., et al. Crack and Noncrack Classification from Concrete Surface Images Using Machine Learning. Structural Health Monitoring, 2019. 18, 725-738 DOI: 10.1177/1475921718768747.
5. Weng, X., Y. Huang, and W. Wang Segment-based pavement crack quantification. Automation in Construction, 2019. 105, DOI: 10.1016/j.autcon.2019.04.014.
6. Gui, R., et al. Object-Based Crack Detection and Attribute Extraction From Laser-Scanning 3D Profile Data. IEEE Access, 2019. 7, 172728-172743 DOI: 10.1109/ACCESS.2019.2956758.
7. Pollock, E. Italy's Morandi Bridge Collapse—What Do We Know? 2018 [cited 2020 20 December 2020]; Available from: https://www.engineering.com/BIM/ArticleID/17517/Italys-Morandi-Bridge-CollapseWhat-Do-We-Know.aspx.
8. Attard, L., et al. Automatic crack detection using mask R-CNN. 2019. IEEE Computer Society.
9. Dorafshan, S., R.J. Thomas, and M. Maguire Comparison of deep convolutional neural networks and edge detectors for image-based crack detection in concrete. Construction and Building Materials, 2018. 186, 1031-1045 DOI: 10.1016/j.conbuildmat.2018.08.011.
10. Computer Vision What it is and why it matters. 28 December 2020]; Available from:

Proceedings of the 10th International Conference on Structural Health Monitoring of Intelligent Infrastructure, SHMII 1011. Liu, Y., J.K.W. Yeoh, and D.K.H. Chua Deep Learning–Based Enhancement of Motion Blurred UAV Concrete Crack Images. Journal of computing in civil engineering, 2020. 34, 4020028 DOI: 10.1061/(ASCE)CP.1943-5487.0000907.
12. Chen, Y., et al. A bridge crack image detection and classification method based On climbing robot. 2016. 4037-4042 DOI: 10.1109/ChiCC.2016.7553984.
13. Sieberth, T., R. Wackrow, and J.H. Chandler Influence of blur on feature matching and a geometric approach for photogrammetric deblurring. International archives of the photogrammetry, remote sensing and spatial information sciences., 2014. XL-3, 321-326 DOI: 10.5194/isprsarchives-XL-3-321-2014.
14. Sieberth, T., R. Wackrow, and J.H. Chandler UAV IMAGE BLUR - ITS INFLUENCE AND WAYS TO CORRECT IT. International archives of the photogrammetry, remote sensing and spatial information sciences., 2015. XL-1-W4, 33-39 DOI: 10.5194/isprsarchives-XL-1-W4-33-2015.
15. Nguyen, A., et al. Deterioration assessment of buildings using an improved hybrid model updating approach and long-term health monitoring data. Structural health monitoring, 2018. 18, 5-19 DOI: 10.1177/1475921718799984.
16. Nguyen, A., et al., Toward effective structural identification of medium-rise building structures. Journal of Civil Structural Health Monitoring, 2018. 8(1): p. 63-75.
17. Gharehbaghi, V.R., et al. Deterioration and damage identification in building structures using a novel feature selection method. in Structures. Elsevier.
18. Gharehbaghi, V.R., et al., Supervised damage and deterioration detection in building structures using an enhanced autoregressive time-series approach. Journal of Building Engineering, 2020. 30: p. 101292.
19. Rosner, M. Transfer Learning & Machine Learning: How It Works, What It's Used For, and Where it's Taking Us. 2018 22 December 2012]; Available from: https://www.sparkcognition.com/transfer-learning-machine-learning/.
20. Bang, S., et al. Encoder–decoder network for pixel-level road crack detection in black-box images. Computer-Aided Civil and Infrastructure Engineering, 2019. 34, 713-727 DOI: 10.1111/mice.12440.
21. Kim, B. and S. Cho, Automated vision-based detection of cracks on concrete surfaces using a deep learning technique. Sensors, 2018. 18(10): p. 3452.
22. Dung, C.V. and L.D. Anh Autonomous concrete crack detection using deep fully convolutional neural network. Automation in Construction, 2019. 99, 52-58 DOI: 10.1016/j.autcon.2018.11.028.
23. Sontakke, M.D., S. Meghana, and M.S. Kulkarni DIFFERENT TYPES OF NOISES IN IMAGES AND NOISE REMOVING TECHNIQUE. 3.
24. How to Control Image Noise. Digital SLR pocket Guide 3rd Edition 2020 [cited 2020 31 August 2020]; Available from: https://www.photoreview.com.au/tips/shooting/how-to-control-image-noise/.
25. Stephens, C. Understanding Image Noise in Your Film and Video Projects. 2018 [cited 2020 31 August 2020]; Available from: https://www.premiumbeat.com/blog/understanding-film-video-image-noise/.
26. Maguire, M., D. S., and T. RJ., SDNET-2018: A concrete crack image dataset for machine learning applications. 2018, Utah State University: Utah State University.
27. Özgenel, C. and F. Çağlar, Concrete Crack Images for Classification, Mendeley, Editor. 2019, Mendeley Data: Mendeley.
28. Concrete Crack. GoogleNet.
29. Lim, J.S., Two-dimensional signal and image processing. ph, 1990.
30. MATLAB Documentation. Available from: https://www.mathworks.com/help/matlab/.
31. Lin, X., et al., Recent advances in passive digital image security forensics: A brief review. Engineering, 2018. 4(1): p. 29-39.
32. Arya, D., et al., Transfer learning-based road damage detection for multiple countries. arXiv preprint arXiv:2008.13101, 2020.
33. Olschofsky, K. and M. Köhl, Rapid field identification of cites timber species by deep learning. Trees, Forests and People, 2020. 2: p. 100016.
34. Bianco, S., et al., Benchmark analysis of representative deep neural network architectures. IEEE Access, 2018. 6: p. 64270-64277.
35. Mathworks. 5 January 2021]; AlexNet]. Available from: https://www.mathworks.com/help/deeplearning/ref/alexnet.html.
36. Han, X., et al., Pre-trained alexnet architecture with pyramid pooling and supervision for high spatial resolution remote sensing image scene classification. Remote Sensing, 2017. 9(8): p. 848.
37. Saxena, S. Precision vs Recall. Precision vs Recall 2018 [cited 2020 21 July 2020]; Available from: https://towardsdatascience.com/precision-vs-recall-386cf9f89488.
8